\documentclass{article}

\usepackage[preprint]{neurips_2023}

\usepackage[utf8]{inputenc} 
\usepackage[T1]{fontenc}    
\usepackage{hyperref}       
\usepackage{url}            
\usepackage{booktabs}       
\usepackage{amsfonts}       
\usepackage{nicefrac}       
\usepackage{microtype}      
\usepackage{xcolor}         
\usepackage{graphicx}
\usepackage{subcaption}
\title{Minor DPO reject penalty to increase training robustness}
\author{%
Shiming Xie \and Hong Chen  \and Fred Yu  \\
\and Zeye Sun \and Xiuyu Wu \and Yingfan Hu  \\
\and  \{shiming.xsm, wuyi.chen, fred.yf, zeye.szy, wuxiuyu.wxy, huyingfan.hyf\}@antgroup.com
}
\date{August 2024}

\begin{document}

\maketitle

\begin{abstract}
  Learning from human preference is a paradigm used in large-scale language model (LLM) fine-tuning step to better align pretrained LLM to human preference for downstream task. In the past it uses reinforcement learning from human feedback (RLHF) algorithm to optimize the LLM policy to align with these preferences and not to draft too far from the original model. Recently, Direct Preference Optimization (DPO) has been proposed to solve the alignment problem with a simplified RL-free method. Using preference pairs of chosen and reject data, DPO models the relative log probability as implicit reward function and optimize LLM policy using a simple binary cross entropy objective directly. DPO is quite straight forward and easy to be understood. It  perform efficiently and well in most cases.  In this article, we analyze the working mechanism of $\beta$ in DPO, disclose its syntax difference   between RL algorithm and DPO,  and understand the potential shortage brought by the DPO simplification. With these insights, we propose MinorDPO, which is better aligned to the original RL algorithm, and increase the stability of preference optimization process.	
\end{abstract}

\section{Background}
Recent LLM trained on very large datasets are extremely powerful in understanding human queries and providing surprisingly reasoning ability. Traditional industry like finance also begin to try using LLM in its business, but require a high demand of accuracy on number, entity, event and etc., which brings in much higher requirements on the preference alignment quality. In the past RLHF establish a paradigm for the alignment work, which use reward model and reinforcement learning like \cite{schulman2017proximal} to optimize the LLM model after a supervised fine-tuning. \cite{rafailov2023direct} present a simplified algorithm DPO which directly optimizes the LLM with a simple cross entropy classification objective based on the preference pair data.  DPO models the relative log probability as an implicit reward function and derived a closed objective form. Its concept is quite straight froward, easy to understand, and appropriate for quick start work. Meanwhile we have also found DPO is fragile in some data distribution, which need a careful tuning to avoid an optimization crash.

DPO update the relative log probability of the preference pair data, using a dynamic sample-level importance weight to prevent the model degeneration. The key factor is the hyper-parameter $\beta$ used in DPO algorithm. \cite{rafailov2023direct} claim $\beta$ accounting for the strength of the KL constraints. In this paper we analyze the internal working mechanism of DPO on how it optimize model, clarify the slightly different syntax of $\beta$ between RL Eq. \ref{ppo_equation} and DPO Eq. \ref{dpo_equation}  , to know the shortage DPO bring in.

Our main contribution is that we give a comprehensive analysis of the $\beta$ mechanism and propose an improved loss function Minor DPO Eq. \ref{minor_dpo_equation} based on the analysis. Minor DPO is better aligned to RL Eq. \ref{ppo_equation} without bringing new hyper-parameters, and it improve the stability and robustness of the preference alignment optimization.

\section{Related Work }
Reinforce Learning from human feedback(\cite{ouyang2022training}, \cite{ziegler2020finetuning}) present a paradigm to align  LLM to human preference. It first creates a reward model from the preference data pair of chosen and reject. Then LLM is trained through RL algorithm like PPO (\cite{schulman2017proximal}) to maximize the learned reward on the LLM while maintaining the discrepancy between the optimized LLM and the original LLM to prevent model degeneration. 

DPO(\cite{rafailov2023direct}) introduce a new parameterization of the reward model in RLHF and allow to solve the standard RLHF problem using a simplified solution with only a cross entropy classification loss based on the preference data pair directly. It's simpler to implement, requires less compute resource, and can fine-tune LLM to align with human as well as RL algorithm.  \cite{rafailov2024r} derive that DPO is token-level MDP that satisfies the Bellman equation and works as a general inverse Q-learning algorithm in a theoretical way. 

There are several other variants that optimize alignment objective based on the preference sample directly. Some variants contain a reference model  while some variants are reference-free.

IPO(\cite{azar2023general}) identify that DPO may be prone to over-fitting in situations where the preference probability of the chosen over the reject sample is close to 1 and propose Identity Preference Optimization based on preference data pair. 

KTO(\cite{ethayarajh2024kto}) propose HALO that directly maximizes the utility of generation, instead of maximizing the log-likelihood of preference. It's able to train with only chosen samples, which is useful in some scenarios.

DPOP(\cite{pal2024smaug}) identify that DPO may lead to degenerate when the preference data pair is close. It uses a theoretic analysis, prove both $reward/chosen$ and $reward/reject$ grows up to negative value with a high probability, and propose a fix by adding an additional non-linear reward on the chosen sample $y_w$, somehow like adding an extra SFT loss conditional on the relative log probability distance of the chosen sample $y_w$ between the optimized LLM and the reference LLM.

PCO(\cite{adolphs2022cringe}) argue that method such as unlikelihood which simply push down the probability of negative tokens may inadvertently push up the probability of low quality or rare tokens for that sequence position. It proposes pairwise cringe loss which use a sampled top-k token, and instead of push down the probability of the original negative token, it increases the probability of the sampled token and thus decrease the original negative token indirectly. The pairwise cringe loss doesn't use a ref model to constraint the discrepancy that the optimized model may draft from the original model.  

SimPO(\cite{meng2024simpo}) propose a simpler approach, it's reference-free and use the average log probability of the sequence as an implicit reward, and introduce a target reward margin to the Bradley-Terry objective.

Orpo (\cite{hong2024orpo}) propose a variant of SFT by using odds ratio-based penalty to the conventional negative log-likelihood (NLL) loss for the reject samples. It's also reference-free.  

Beyond the RL and DPO variants, \cite{nakano2022webgpt}, \cite{askell2021general} \cite{cobbe2021training} explore best-of-n sampling to improve large language model generation by selecting the best response based on the human preference rewards among n sampled responses. RRHF(\cite{yuan2023rrhf}) is targeted to learn the best response and comparisons based on the human preference rewards among n sampled responses to achieve alignment during optimization instead of inference.

\section{Approach}
\subsection{DPO derivation }
\cite{ziegler2020finetuning} presents the RLHF pipeline paradigm. In the RL Fine-tuning phase, it uses the reward model learned in reward modeling phase to optimize the LLM model to retrieve maximum reward using a classical RL objective.

\begin{equation}
	\max_x{\mathbb E_{x \sim D, y \sim \pi_\theta(y|x)}[r_\phi(x,y)] - \beta\mathbb D_{kl}[\pi_\theta(y|x) || \pi_{ref}(y|x)]} \label{ppo_equation}
\end{equation}
$r_\phi(x,y)$ here is the reward function. $\mathbb D_{kl}[\pi_\theta(y|x) || \pi_{ref}(y|x)]$  is  a bound that constraint $\pi_\theta(y|x) $ not to deviate from $\pi_{ref}(y|x)$. And hyper-parameter $\beta$ presents the constraints strength. The KL constraints is important as it prevent the optimized model over-optimized on the training datasets, which may cause the model degenerate and lose the generalization ability learned during the pre-train phase. Due to the nature of discrete language generation, the objective is optimized with RL algorithm using PPO(\cite{schulman2017proximal}) or PPO with additional pre-train loss (ppo-ptx in \cite{ouyang2022training}).

Due to the computation complexity of RL, \cite{rafailov2023direct} propose a simplified implementation and derive  DPO objective from RL in a closed form.

\begin{equation}
	L_{DPO}(\pi_\theta;\pi_{ref}) 
	=-\mathbb E_{(x,y_w,y_l) \sim D}[{log\sigma(\beta log \frac{\pi_\theta(y_w|x)}{\pi_{ref}(y_w|x)} - \beta  log \frac{\pi_\theta(y_l|x)}{\pi_{ref}(y_l|x)})} ] \label{dpo_equation}
\end{equation}

The DPO objective optimize the model directly using a simple cross entropy loss based on the preference pair data. It avoids the reward model, critic model and sampling process used in the RL algorithm. 

let's expand DPO loss equation and get DPO gradient equation.

\begin{equation}
	\nabla_\theta L_{DPO}(\pi_\theta;\pi_{ref}) \\
	=-\beta\mathbb E_{(x,y_w,y_l) \sim D}[\sigma(-\beta log \frac{\pi_\theta(y_w|x)}{\pi_{ref}(y_w|x)} + \beta  log \frac{\pi_\theta(y_l|x)}{\pi_{ref}(y_l|x)})[\nabla_\theta log\pi_\theta(y_w|x) - \nabla_\theta log\pi_\theta(y_l|x)] ] \label{dpo_gradient}
\end{equation}

In this article we use $\textbf{rewards/chosen}$ to represent $log \frac{\pi_\theta(y_w|x)}{\pi_{ref}(y_w|x)}$, use $\textbf{rewards/reject}$\footnote{We may use chosen, reject for short in this article} to represent $log \frac{\pi_\theta(y_l|x)}{\pi_{ref}(y_l|x)}) $, and use $\textbf{margin}$ to represent $log \frac{\pi_\theta(y_w|x)}{\pi_{ref}(y_w|x)} - log \frac{\pi_\theta(y_l|x)}{\pi_{ref}(y_l|x)}$.

$rewards/chosen$ is how much the optimized model prefer the chosen sample in preference data pair over the reference model.   $rewards/reject$ is how much the optimized model prefer the reject sample in preference data pair over the reference model.  And $margin$ is how much the optimized model prefer the chosen sample, and dis-prefer the reject sample in preference data pair over the reference model. In a summary,  DPO objective is to enlarge the margin between the chosen samples and the reject samples.

And thus Eq. \ref{dpo_gradient}, can be rewritten into

\begin{equation}
	\nabla_\theta L_{DPO}(\pi_\theta;\pi_{ref}) 
	=-\mathbb E_{(x,y_w,y_l) \sim D}[ \beta * \sigma(-\beta * margin)[\nabla_\theta log\pi(y_w|x) - \nabla_\theta log\pi(y_l|x)] ] \label{dpo_gradient_abbr}
\end{equation}

The loss gradient can be divided into three parts, 
\begin{enumerate}
	\item A  sample level dynamic coefficient $f(\beta, margin) = \beta * \sigma(-\beta * margin)$ related to $\beta$ and $margin$
	\item A reward  $\nabla_\theta log\pi(y_w|x)$ to the positive sample $y_w$
	\item A penalty $- \nabla_\theta log\pi(y_l|x) $ to the reject sample $y_l$.
\end{enumerate}

\subsection{ Different \texorpdfstring{$\beta$}\  syntax  between DPO and RL}
In \cite{rafailov2023direct} it claims $\beta$ account for the strength of the KL constraints, something same as the $\beta$ hyper-parameter used in RL algorithm. Let's dig into the sample level dynamic coefficient and see how $\beta$ affect the training process.

\begin{figure}[htbp]
	\centering
	\begin{minipage}{0.75\textwidth}
		\includegraphics[width=\textwidth]{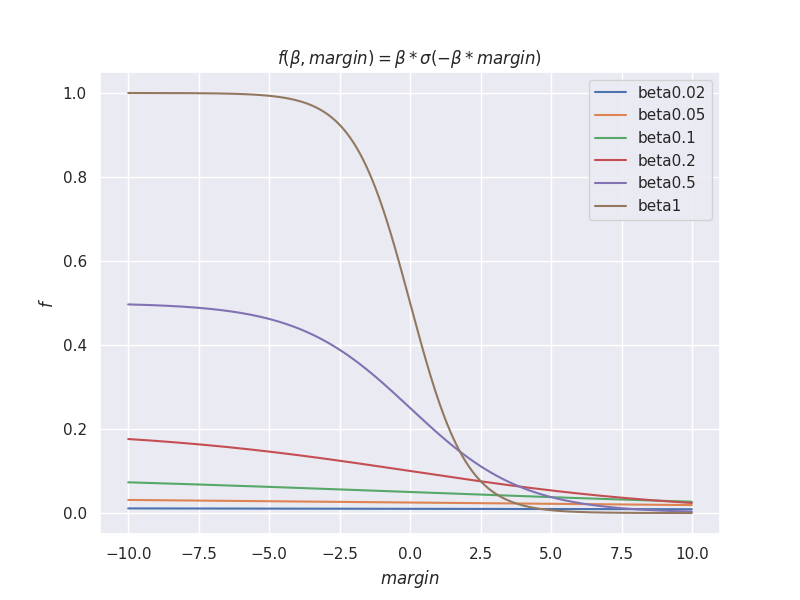} 
	\end{minipage}
	\caption{sample level dynamic coefficient $f(\beta, margin)$ comparison}
	\label{beta_comparation}
\end{figure}

Figure \ref{beta_comparation} shows that $\beta$ control the shape of the coefficient $f(\beta, margin)$ . The bigger the $\beta$ is , the sooner the $f(\beta, margin)$ decay, especially when the margin is positive(As  DPO is to enlarge the margin, so the margin shall grow to positive ). During the training process, when the margin grows up to some threshold value, bigger $\beta$ will return a smaller coefficient than smaller $\beta$, and affect the gradient magnitude in the back propagation process. In summary,
\begin{enumerate}
	\item For each $\beta$, low margin value give a high coefficient value.
	\item Efficient $f(\beta, margin)$ with bigger $\beta$  decay faster.
	\item $\beta$ affect the training process through the dynamic coefficient, and alter the update magnitude by interfere with the learning rate indirectly . 
\end{enumerate}

Here we present some trends diagram of $rewards/chosen$, $rewards/reject$ and $rewards/margin$ during training, to get some intuitive understandings.

\begin{figure}[htbp]
	\centering
	\begin{minipage}{0.22\textwidth}
		\includegraphics[width=\textwidth]{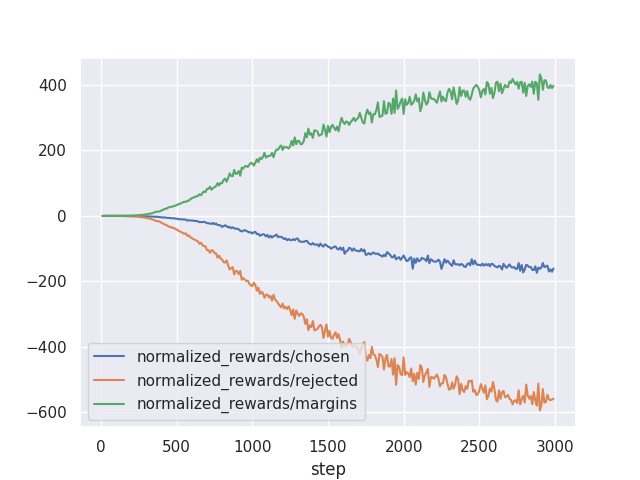} 
		\subcaption{$\beta$=0.02}
	\end{minipage}
	\begin{minipage}{0.22\textwidth}
		\includegraphics[width=\textwidth]{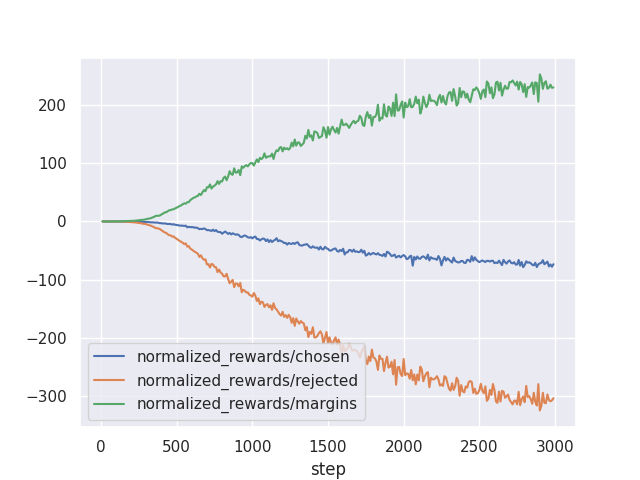} 
		\subcaption{$\beta$=0.04}
	\end{minipage}
	\begin{minipage}{0.22\textwidth}
		\includegraphics[width=\textwidth]{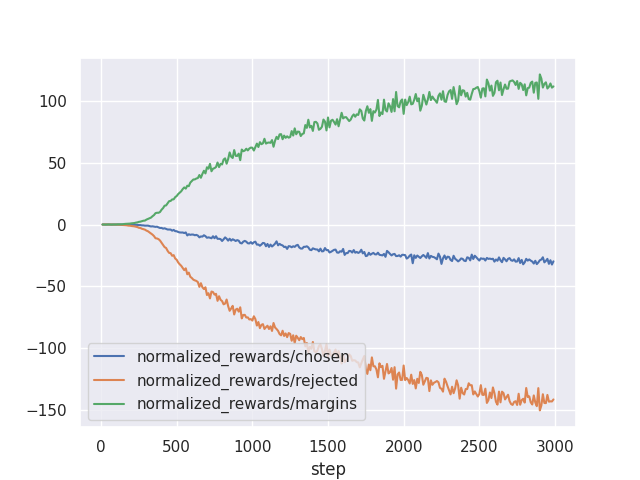} 
		\subcaption{$\beta$=0.1}
	\end{minipage}
	\begin{minipage}{0.22\textwidth}
		\includegraphics[width=\textwidth]{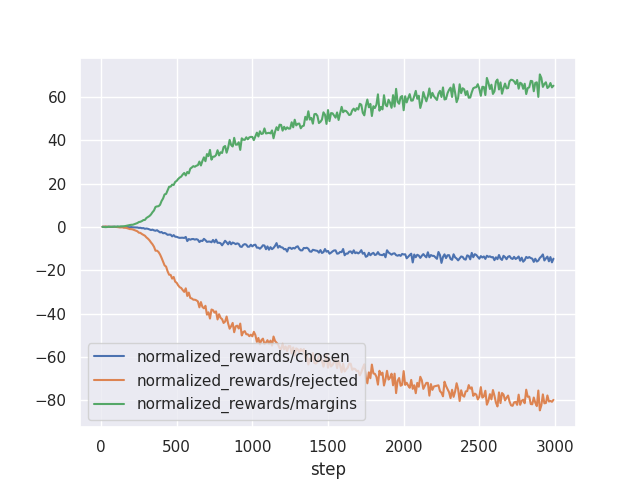} 
		\subcaption{$\beta$=0.2}
	\end{minipage}
	\caption[caption for comparison]{DPO  rewards comparison\footnotemark  with $\beta$, learning\_rate=$1e-6$ }
	\label{reward_compare_1em6}	
\end{figure}

\footnotetext{For those rewards comparison figures, we use normalized\_rewards to mean it is $\beta$-free so that rewards values can be compared between different $\beta$.}

Figure \ref{reward_compare_1em6} shows a real training process for $\beta$ value between 0.02 and 0.2 with learning rate $1e-6$. Each sub-figure shows the $rewards/margin$ grows up during the process, as expected. However, $rewards/chosen$ and $rewards/reject$ both grow to negative, which is a surprise and we will come back to it in \ref{dpo_shortage}. And between those sub-figures, we find that
\begin{enumerate}
	\item Higher $\beta$ have a lower $rewards/margin$ value
	\item Higher $\beta$ also have a lower absolute value of $rewards/chosen$ and $rewards/reject$
\end{enumerate}

\begin{figure}[htbp]
	\centering
	\begin{minipage}{0.22\textwidth}
		\includegraphics[width=\textwidth]{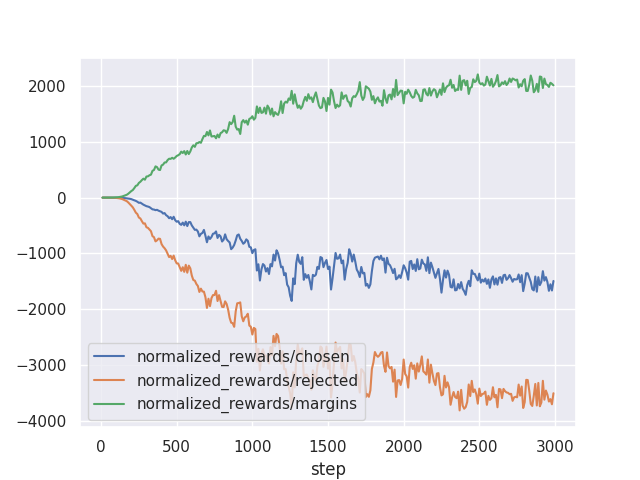} 
		\subcaption{$\beta$=0.02}
	\end{minipage}
	\begin{minipage}{0.22\textwidth}
		\includegraphics[width=\textwidth]{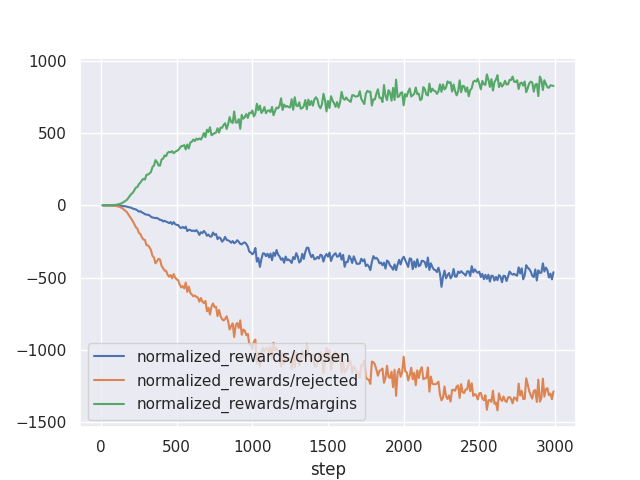} 
		\subcaption{$\beta$=0.04}
	\end{minipage}
	\begin{minipage}{0.22\textwidth}
		\includegraphics[width=\textwidth]{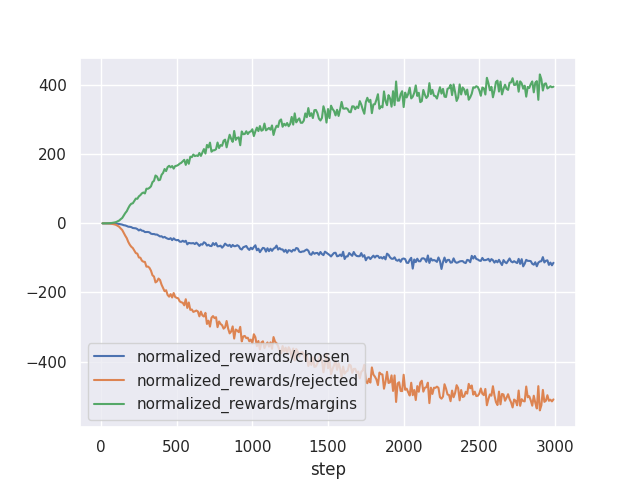} 
		\subcaption{$\beta$=0.1}
	\end{minipage}
	\begin{minipage}{0.22\textwidth}
		\includegraphics[width=\textwidth]{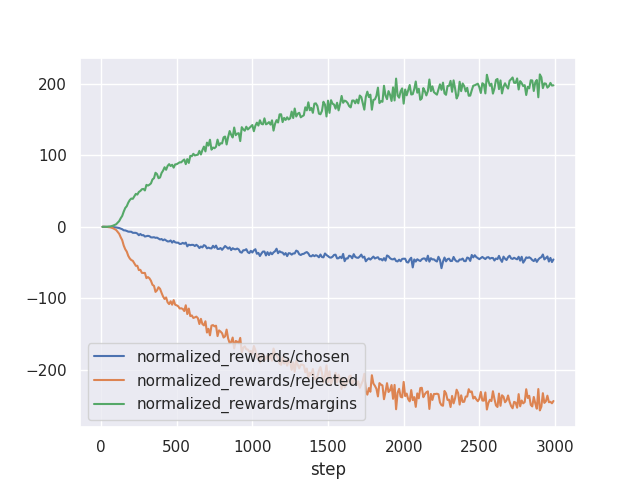} 
		\subcaption{$\beta$=0.2}
	\end{minipage}
	\caption{DPO rewards comparison with $\beta$, learning\_rate=$1e-5$}
	\label{reward_compare_1em5}	
\end{figure}
As said $\beta$ control the shape of coefficient $f(\beta, margin)$, which  finally affect the training process, interleaving with the learning rate. Figure \ref{reward_compare_1em5} shows another comparison for $\beta$ value between 0.02 and 0.2 with learning rate 1e-5. Sub-figures in figure \ref{reward_compare_1em5} have same trends as in figure \ref{reward_compare_1em6}. And comparison of same $\beta$ between figure \ref{reward_compare_1em6} and figure \ref{reward_compare_1em5} shows that the bigger learning rate will have bigger margin. 

It shows that $\beta$ represent certain constraints strength by interleaving with the learning rate for each training sample. During the training process, when the margin grows up to certain value, the coefficient $f(\beta, margin)$ will return a really small value, which interleave with the learning rate and result a really small update step. So that $\beta$ also have an impact on the $rewards/chosen$ and $rewards/reject$, which can be verified through above figures.

But the surprise and import things here is that $beta$ somehow is more related to the $rewards/margin$, instead of the $rewards/chosen$ and $rewards/reject$.  DPO $\beta$ is a constraints strength on the margin value.  The DPO loss gradient descent process is to enlarge the margin, and doesn't control the KL value $rewards/chosen$ and $rewards/reject$ directly as in the RL Eq. \ref{ppo_equation}. 

\textbf{Syntax difference:} This is the fundamental different syntax of $\beta$ between RL and DPO. In DPO $\beta$ is only a constraints strength on the $margin$, but not a direct constraint to $rewards/choosn$ and $rewards/reject$.  So, it may cause both $rewards/chosen$ and $rewards/reject$ grow negative as long as the margin is still growing up, as showed in the figure. It causes some unwanted result such as model degeneration stated below.  

\subsection{DPO shortage}
\label{dpo_shortage}
In \cite{pal2024smaug} the author has a theoretic analysis on why both $rewards/chosen$ and $rewards/reject$ grows up to negative when the preference samples are close, and it propose a fix by adding an additional non-linear reward on the $y_w$ sample.

\begin{equation}
	L_{DPOP}(\pi_\theta;\pi_{ref}) 
	=-\mathbb E_{(x,y_w,y_l) \sim D}[log\sigma(\beta log \frac{\pi_\theta(y_w|x)}{\pi_{ref}(y_w|x)} - \beta  log \frac{\pi_\theta(y_l|x)}{\pi_{ref}(y_l|x)} - \beta \lambda max(0, log \frac{\pi_{ref}(y_w|x)}{\pi_\theta(y_w|x)})) ] \label{dpop_loss}
\end{equation}

It's same as adding an additional SFT loss on  $y_w$, but somehow use a non-linear transformation with  $\lambda max(0, log \frac{\pi_{ref}(y_w|x)}{\pi_\theta(y_w|x)})$ format. It adds  SFT loss  only when  $\pi_\theta(y_w|x)$ is smaller than $\pi_{ref}(y_w|x)$ so that  $log \frac{\pi_{ref}(y_w|x)}{\pi_\theta(y_w|x)} \ge 0$ . It works like a compensation for the over penalty to the reject samples. The additional loss shall bring $rewards/chosen$ $log \frac {\pi_\theta(y_w|x)}{\pi_{ref}(y_w|x)} \ge 0 $  for some data distribution. while it also brings an additional hyper parameter $\lambda$ that needs to be tuned with the corresponding $\beta$ value. 

Besides $rewards/chosen$ grows up to negative, in \cite{adolphs2022cringe}, \cite{jiang2022simple} they argue that method such as unlikelihood which simply push down the probability of reject tokens may inadvertently push up the probability of low quality or rare tokens for that sequence position, because there is no control over that effect. We have observed the same problem in our experiments, with a slightly larger learning rate like $5e-6$ to $1e-5$, the DPO optimized model may start to generate repeated token, or some unwanted symbol token in the answer, which we believe is caused by the over penalty.  Although with an appropriate small learning rate, the DPO optimized model works, we still believe if they're enough training steps, DPO may still crash the model when the accumulated penalty on the reject samples surpasses a certain threshold. 

In \cite{adolphs2022cringe} it uses a sampled token and instead of pushing down the probability of the original reject token, it increases the probability of the sampled token. Compared to the standard cringe loss, it uses the preference pair and get a dynamic margin between $logp(y_w|x)$ and $logp(y_l|x)$ to control the pair cringe loss. This method brings several hyper-parameters to tune, and in our experiment it may over-pull up the sampled token probability and cause the optimized model generate repeat token. 

\subsection{Minor DPO reject penalty to improve training robustness }
They're several assumptions that we found really important and the our solution is based on those assumptions.
\begin{enumerate}
	\item the Eq.\ref{ppo_equation} suggest that the $\pi_\theta$ should not far from $\pi_{ref}$, for both $y_w$ and $y_l$, and it use an explicit KL loss to present the constraints, and use $\beta$ to control the constraints strength.
	
	\item $\pi_{ref}$ in DPO is quite important and is the key factor to prevent over optimization.  DPO use $log \frac{\pi_\theta}{\pi_{ref}}$ to represent the dispersion between the optimized $\pi_\theta$ and $\pi_{ref}$,  and in theoretical we expect $log \frac{\pi_\theta(y_w|x)}{\pi_{ref}(y_w|x)} \ge 0$ and $log \frac{\pi_\theta(y_l|x)}{\pi_{ref}(y_l|x)} \le 0$  for the optimized model.
	
	\item Although \cite{rafailov2023direct} claims $\beta$ in DPO accounting for the strength of the KL constraints,  with above analysis, $\beta$ is the constraints strength for margin and affect the training process by interleaving with the learning rate, instead of  affecting  $\pi_\theta$ and $\pi_{ref}$ directly as in Eq. \ref{ppo_equation}
	
	\item  \cite{pal2024smaug}  proves that DPO will lead both $\pi_\theta(y_w|x) \le \pi_{ref}(y_w|x)$ and $\pi_\theta(y_l|x) \le \pi_{ref}(y_l|x)$ when $y_w$ and $y_l$ is close, and  \cite{adolphs2022cringe} declare that simply push down the probability of reject tokens may lead the optimized model to crash(actually in this case it's even worse as both $\pi_\theta(y_w|x) \le \pi_{ref}(y_w|x)$ and $\pi_\theta(y_l|x) \le \pi_{ref}(y_l|x)$ )
\end{enumerate}

Thus, we propose Minor DPO reject penalty equation

\begin{equation}
	L_{MinorDPO}(\pi_\theta;\pi_{ref}) \\
	=-E_{(x,y_w,y_l) \sim D} {log\sigma(\beta log \frac{\pi_\theta(y_w|x)}{\pi_{ref}(y_w|x)} - \beta max(0, log \frac{\pi_\theta(y_l|x)}{\pi_{ref}(y_l|x)}))}   \label{minor_dpo_equation}
\end{equation}

The Eq.\ref{minor_dpo_equation} adds an additional constraints to reject samples, by replacing the original penalty $-log\frac{\pi_\theta(y_l|x)}{\pi_{ref}(y_l|x)}$ with $-max(0, log\frac{\pi_\theta(y_l|x)}{\pi_{ref}(y_l|x)})$. It decreases penalty on  reject samples, and hence we name it Minor DPO reject penalty and MinorDPO in short.

\begin{figure}[htbp]
	\centering
	\begin{minipage}{0.22\textwidth}
		\includegraphics[width=\textwidth]{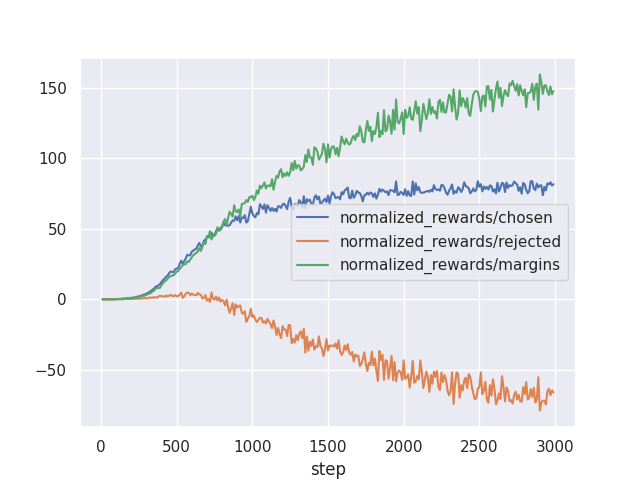} 
		\subcaption{$\beta$=0.02}
	\end{minipage}
	\begin{minipage}{0.22\textwidth}
		\includegraphics[width=\textwidth]{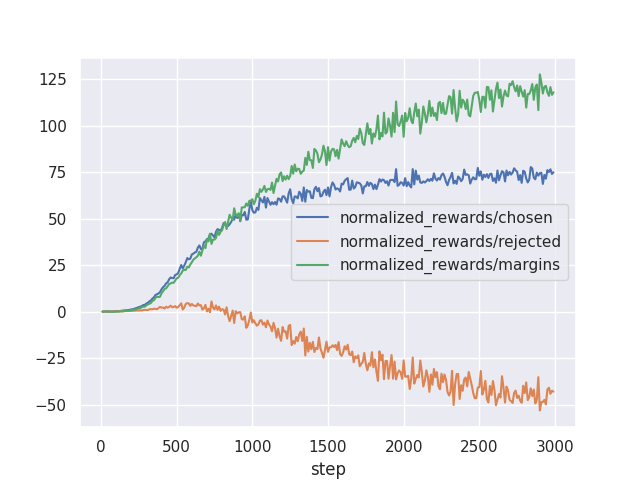} 
		\subcaption{$\beta$=0.04}
	\end{minipage}
	\begin{minipage}{0.22\textwidth}
		\includegraphics[width=\textwidth]{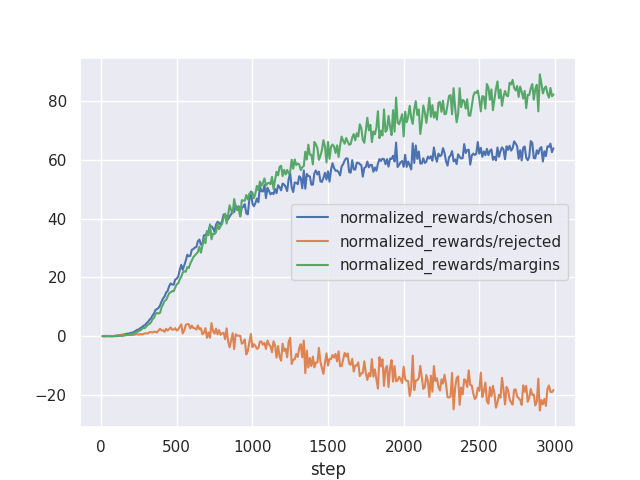} 
		\subcaption{$\beta$=0.1}
	\end{minipage}
	\begin{minipage}{0.22\textwidth}
		\includegraphics[width=\textwidth]{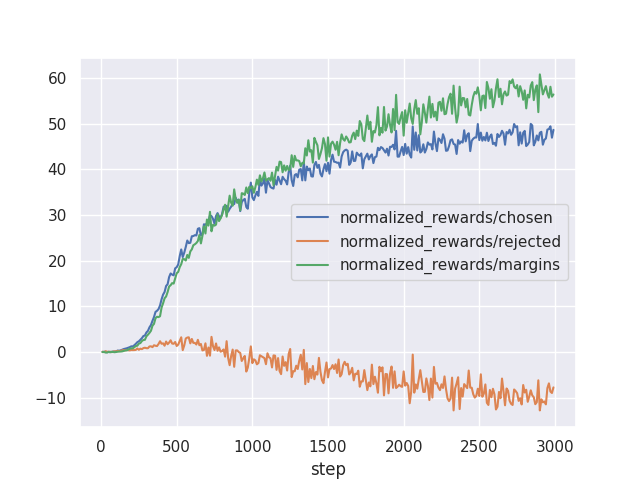} 
		\subcaption{$\beta$=0.2}
	\end{minipage}
	\caption{MinorDPO rewards comparison with $\beta$, learning\_rate=$1e-6$}
	\label{minor_dpo_reward_compare_1em6}	
\end{figure}

\begin{figure}[htbp]
	\centering
	\begin{minipage}{0.22\textwidth}
		\includegraphics[width=\textwidth]{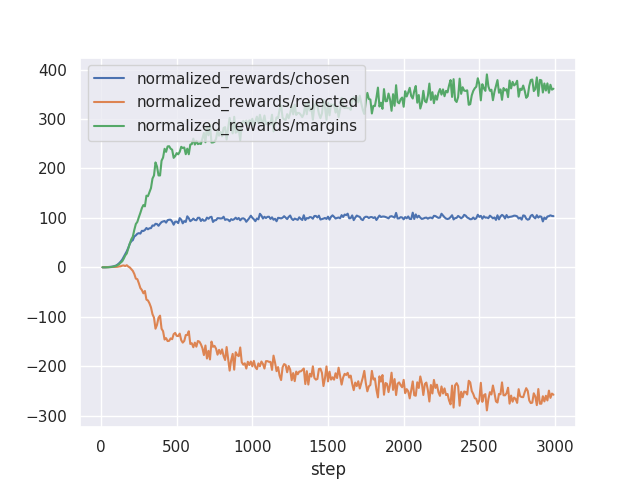} 
		\subcaption{$\beta$=0.02}
	\end{minipage}
	\begin{minipage}{0.22\textwidth}
		\includegraphics[width=\textwidth]{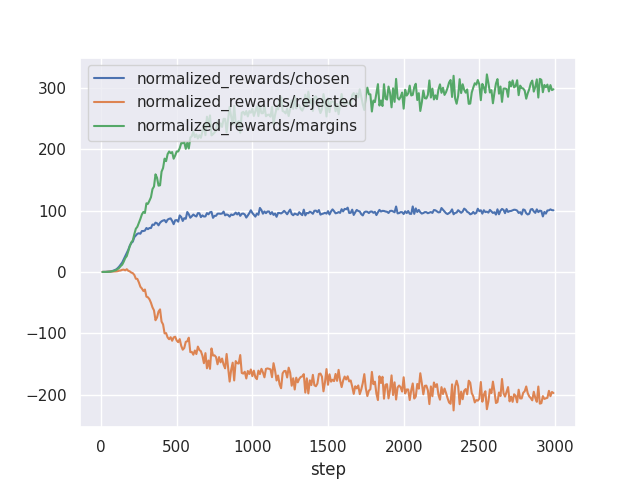} 
		\subcaption{$\beta$=0.04}
	\end{minipage}
	\begin{minipage}{0.22\textwidth}
		\includegraphics[width=\textwidth]{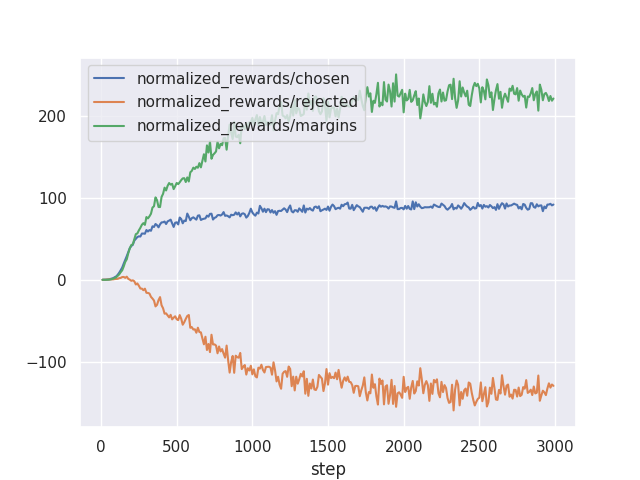} 
		\subcaption{$\beta$=0.1}
	\end{minipage}
	\begin{minipage}{0.22\textwidth}
		\includegraphics[width=\textwidth]{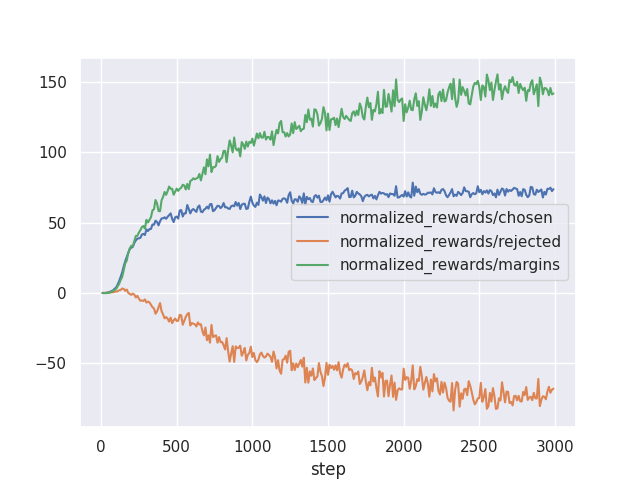} 
		\subcaption{$\beta$=0.2}
	\end{minipage}
	\caption{MinorDPO rewards comparison with $\beta$, learning\_rate=$1e-5$}
	\label{minor_dpo_reward_compare_1em5}	
\end{figure}

Figure \ref{minor_dpo_reward_compare_1em6} and \ref{minor_dpo_reward_compare_1em5} shows that  $rewards/chosen$ grows up to positive during the training process on same datasets used in DPO, and the magnitude of $rewards/chosen$, $rewards/reject$, $rewards/margin$ is much more reasonable. 

Minor DPO is closer to original RL target Eq. \ref{ppo_equation} by adding an explicit constraint on the reject penalty. 
\begin{enumerate}
	\item For those reject samples, $\pi_\theta$ is more close to $\pi_{ref}$ as we stop the gradient descent on rejects samples that are $\pi_\theta(y_l|x) \le \pi_{ref}(y_l|x)$, as we don't want to over-optimized on the rejected samples
	\item For those positive samples, we could achieve $\pi_\theta(y_w|x) \ge \pi_{ref}(y_w|x)$
\end{enumerate}

Comparing to DPO, MinorDPO decreases the reject penalty and slightly increases chosen reward for samples that are $\pi_\theta(y_l|x) \le \pi_{ref}(y_l|x)$ . It ease the over-penalty problem and doesn't introduce  new hyper-parameter. In most cases, it can use the same DPO hyper-parameter settings(or a bigger learning rate for training).  We find that minor DPO is much more robust and can accept a much higher learning rate without crashing the model. (A note here is that with higher learning rate, bigger batch size is needed to maintain the training stability)

\section{Experiments}
Since DPO, there exists several DPO-variants that propose some modifications for different purpose. Here we bring DPOP in the comparison because DPOP is the only methods we know so far that also notice DPO will degenerate when the preference training data is close. DPOP propose an addition SFT on the chosen sample , while we believe the degeneration is caused by the over penalty on the reject samples from DPO, and propose a solution that decrease the penalty, to make it more consistent with the original RL method.

For training settings, we use Qwen1.5-7B-Chat(\cite{qwen}) as the base model, use MetaMath(\cite{yu2024metamath}) as training data with the way mentioned in \cite{pal2024smaug}\footnote{use data in \href{https://huggingface.co/datasets/abacusai/MetaMath\_DPO\_FewShot}{https://huggingface.co/datasets/abacusai/MetaMath\_DPO\_FewShot} , but change few-shot to zero-shot for training  }, and use test set of GSM8K(\cite{cobbe2021training}) to compare. 

We use LLaMa-Factory(\cite{zheng2024llamafactory}) as the training and inference framework with some customized code to implement the DPOP and MinorDPO algorithm. The experiments use batch size 128, warm-up ratio 0.1, linear decay learning rate, 1 epoch and run 3000+ steps. 

As DPOP has an additional hyper-parameter $\lambda$ to control the compensation strength we use $\lambda$ = 50, which is proposed in \cite{pal2024smaug}.

For  inference settings, we use the prompt  \textit{You are a helpful assistant. Please answer follow question step by step and give final answer in a separate line using the brief format So the answer is:  . Below is the question: } to concat the question content.

\begin{figure}[htbp]
	\centering
	\begin{minipage}{0.45\textwidth}
		\includegraphics[width=\textwidth]{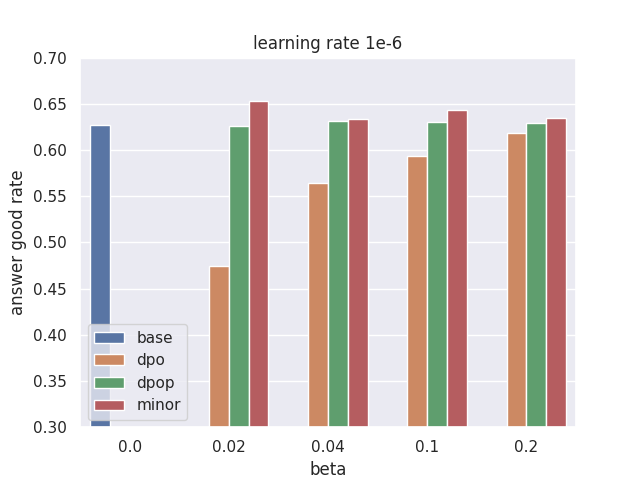} 
		\subcaption{$lr$=1e-6}
		\label{gsm8k_score_1e6}
	\end{minipage}
	\begin{minipage}{0.45\textwidth}
		\includegraphics[width=\textwidth]{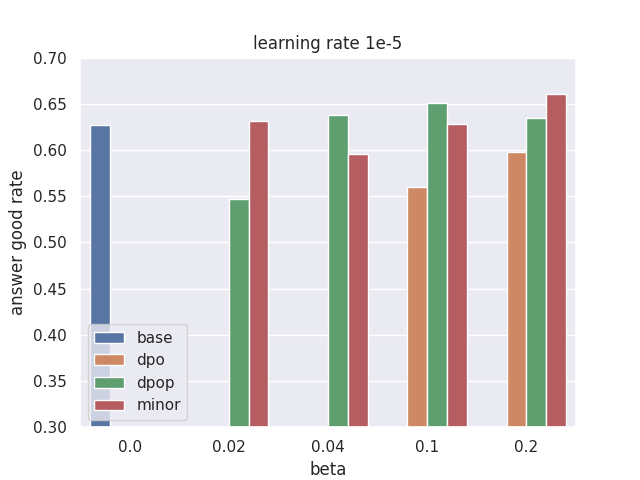} 
		\subcaption{$lr$=1e-5 \footnotemark}
		\label{gsm8k_score_1e5}
	\end{minipage}
	\caption{gsm8k score comparison DPO/DPOP/MinorDPO}
	\label{gsm8k_score}	
\end{figure}
\footnotetext{when learning rate is 1e-5, and $\beta=0.02, 0.04$, the DPO model degenerate seriously and generate repeat tokens, so there are no evaluation result for these two settings}

Figure \ref{gsm8k_score} show the experiment result. We also evaluate the base model and show it in both sub-figures for the comparison. It doesn't need training so its $\beta$ value is 0.  Figure \ref{gsm8k_score} shows several important points:
\begin{enumerate}
	\item Due the closeness of the preference data pair, DPO lost in all settings, MinorDPO wins in most settings, and gets the top score for both learning rate of $1e-5$ and $1e-6$.
	\item DPO performance decrease when $\beta$ become small from 0.2 to 0.02.  when $lr=le-5$ and $\beta$=0.02, 0.04, the result model will generate repeated token in it answer.
	\item DPOP and MinorDPO accept high learning rate, but when $lr=1e-5$, $\beta$=0.02, DPOP will also generate repeat token occasionally, and in Figure \ref{gsm8k_score_1e5} shows it decrease performance seriously, which indicate that the hyper-parameter $\lambda$ is also highly related to learning rate and $\beta$ in order to compensate the penalty.
\end{enumerate}

Related code, data and evaluation output will open source later.

\section{Conclusion \& Future work}
In this article we analyze the different syntax of $\beta$ between DPO and RL, and find DPO shortage it brings in. For the preference sample pair, normally we just prefer the chosen samples over the reject samples, but it doesn't mean the reject samples are totally wrong. The symmetric form of increase on the chosen sample and decrease on the reject sample introduce too much penalty over the reject sample. DPO uses a sample level dynamic coefficient to ease the problem but doesn't solve it completely. We propose  MinorDPO that break the symmetry by reducing the decrease on the reject samples.  In this article we only focus on the over penalty problem thus only compare three methods: DPO, DPOP and MinorDPO using the MetaMath datasets. There may be more methods and more datasets to be tested in the future. We believe DPO is an elegant and efficient base to start with and investigate in for the preference alignment work. MinorDPO is an improvement on DPO and doesn't bring in additional hyper-parameter. With virtually no tuning of hyper-parameter, it performs same or better than DPO and other DPO variants. 

As in our test, MinorDPO can accept higher learning rate. MinorDPO decrease the penalty over reject samples and thus may introduce under-fit problem for some data distributions, so an appropriate higher learning rate may help fixing it.

Another important work is how to prevent over-fit on the learning datasets. We know DPO and DPO variants are sensitive to the $\beta$ and learning rate, so tuning these two hyper-parameters may help. Inspired by $\beta$-free rewards and margin, We think there may exist some training metrics that can be used to quantify whether training is sufficient or not, and thus those metrics can be used inside DPO/MinorDPO in some way to solve the over-fit problem.

\bibliography{main}

\begin{thebibliography}{}

\bibitem[Adolphs et~al., 2022]{adolphs2022cringe}
Adolphs, L., Gao, T., Xu, J., Shuster, K., Sukhbaatar, S., and Weston, J.
  (2022).
\newblock The cringe loss: Learning what language not to model.

\bibitem[Askell et~al., 2021]{askell2021general}
Askell, A., Bai, Y., Chen, A., Drain, D., Ganguli, D., Henighan, T., Jones, A.,
  Joseph, N., Mann, B., DasSarma, N., Elhage, N., Hatfield-Dodds, Z.,
  Hernandez, D., Kernion, J., Ndousse, K., Olsson, C., Amodei, D., Brown, T.,
  Clark, J., McCandlish, S., Olah, C., and Kaplan, J. (2021).
\newblock A general language assistant as a laboratory for alignment.

\bibitem[Azar et~al., 2023]{azar2023general}
Azar, M.~G., Rowland, M., Piot, B., Guo, D., Calandriello, D., Valko, M., and
  Munos, R. (2023).
\newblock A general theoretical paradigm to understand learning from human
  preferences.

\bibitem[Bai et~al., 2023]{qwen}
Bai, J., Bai, S., Chu, Y., Cui, Z., Dang, K., Deng, X., Fan, Y., Ge, W., Han,
  Y., Huang, F., Hui, B., Ji, L., Li, M., Lin, J., Lin, R., Liu, D., Liu, G.,
  Lu, C., Lu, K., Ma, J., Men, R., Ren, X., Ren, X., Tan, C., Tan, S., Tu, J.,
  Wang, P., Wang, S., Wang, W., Wu, S., Xu, B., Xu, J., Yang, A., Yang, H.,
  Yang, J., Yang, S., Yao, Y., Yu, B., Yuan, H., Yuan, Z., Zhang, J., Zhang,
  X., Zhang, Y., Zhang, Z., Zhou, C., Zhou, J., Zhou, X., and Zhu, T. (2023).
\newblock Qwen technical report.
\newblock {\em arXiv preprint arXiv:2309.16609}.

\bibitem[Cobbe et~al., 2021]{cobbe2021training}
Cobbe, K., Kosaraju, V., Bavarian, M., Chen, M., Jun, H., Kaiser, L., Plappert,
  M., Tworek, J., Hilton, J., Nakano, R., Hesse, C., and Schulman, J. (2021).
\newblock Training verifiers to solve math word problems.

\bibitem[Ethayarajh et~al., 2024]{ethayarajh2024kto}
Ethayarajh, K., Xu, W., Muennighoff, N., Jurafsky, D., and Kiela, D. (2024).
\newblock Kto: Model alignment as prospect theoretic optimization.

\bibitem[Hong et~al., 2024]{hong2024orpo}
Hong, J., Lee, N., and Thorne, J. (2024).
\newblock Orpo: Monolithic preference optimization without reference model.

\bibitem[Jiang et~al., 2022]{jiang2022simple}
Jiang, S., Zhang, R., Vakulenko, S., and de~Rijke, M. (2022).
\newblock A simple contrastive learning objective for alleviating neural text
  degeneration.

\bibitem[Meng et~al., 2024]{meng2024simpo}
Meng, Y., Xia, M., and Chen, D. (2024).
\newblock Simpo: Simple preference optimization with a reference-free reward.

\bibitem[Nakano et~al., 2022]{nakano2022webgpt}
Nakano, R., Hilton, J., Balaji, S., Wu, J., Ouyang, L., Kim, C., Hesse, C.,
  Jain, S., Kosaraju, V., Saunders, W., Jiang, X., Cobbe, K., Eloundou, T.,
  Krueger, G., Button, K., Knight, M., Chess, B., and Schulman, J. (2022).
\newblock Webgpt: Browser-assisted question-answering with human feedback.

\bibitem[Ouyang et~al., 2022]{ouyang2022training}
Ouyang, L., Wu, J., Jiang, X., Almeida, D., Wainwright, C.~L., Mishkin, P.,
  Zhang, C., Agarwal, S., Slama, K., Ray, A., Schulman, J., Hilton, J., Kelton,
  F., Miller, L., Simens, M., Askell, A., Welinder, P., Christiano, P., Leike,
  J., and Lowe, R. (2022).
\newblock Training language models to follow instructions with human feedback.

\bibitem[Pal et~al., 2024]{pal2024smaug}
Pal, A., Karkhanis, D., Dooley, S., Roberts, M., Naidu, S., and White, C.
  (2024).
\newblock Smaug: Fixing failure modes of preference optimisation with
  dpo-positive.

\bibitem[Rafailov et~al., 2024]{rafailov2024r}
Rafailov, R., Hejna, J., Park, R., and Finn, C. (2024).
\newblock From $r$ to $q^*$: Your language model is secretly a q-function.

\bibitem[Rafailov et~al., 2023]{rafailov2023direct}
Rafailov, R., Sharma, A., Mitchell, E., Ermon, S., Manning, C.~D., and Finn, C.
  (2023).
\newblock Direct preference optimization: Your language model is secretly a
  reward model.

\bibitem[Schulman et~al., 2017]{schulman2017proximal}
Schulman, J., Wolski, F., Dhariwal, P., Radford, A., and Klimov, O. (2017).
\newblock Proximal policy optimization algorithms.

\bibitem[Yu et~al., 2024]{yu2024metamath}
Yu, L., Jiang, W., Shi, H., Yu, J., Liu, Z., Zhang, Y., Kwok, J.~T., Li, Z.,
  Weller, A., and Liu, W. (2024).
\newblock Metamath: Bootstrap your own mathematical questions for large
  language models.

\bibitem[Yuan et~al., 2023]{yuan2023rrhf}
Yuan, Z., Yuan, H., Tan, C., Wang, W., Huang, S., and Huang, F. (2023).
\newblock Rrhf: Rank responses to align language models with human feedback
  without tears.

\bibitem[Zheng et~al., 2024]{zheng2024llamafactory}
Zheng, Y., Zhang, R., Zhang, J., Ye, Y., Luo, Z., and Ma, Y. (2024).
\newblock Llamafactory: Unified efficient fine-tuning of 100+ language models.
\newblock {\em arXiv preprint arXiv:2403.13372}.

\bibitem[Ziegler et~al., 2020]{ziegler2020finetuning}
Ziegler, D.~M., Stiennon, N., Wu, J., Brown, T.~B., Radford, A., Amodei, D.,
  Christiano, P., and Irving, G. (2020).
\newblock Fine-tuning language models from human preferences.

\end{thebibliography}

\end{document}